\documentclass[10pt,twocolumn,letterpaper]{article}

\usepackage{cvpr}
\usepackage{times}
\usepackage{epsfig}
\usepackage{graphicx}
\usepackage{amsmath}
\usepackage{amssymb}

\usepackage[breaklinks=true,bookmarks=false]{hyperref}

\cvprfinalcopy % *** Uncomment this line for the final submission

\def\cvprPaperID{****} % *** Enter the CVPR Paper ID here

\begin{document}

%%%%%%%%% TITLE
\title{When Ensembling Smaller Models is More Efficient than Single Large Models}

\author{Dan Kondratyuk, Mingxing Tan, Matthew Brown, Boqing Gong\\
Google AI\\
% Institution1 address\\
{\tt\small \{dankondratyuk,tanmingxing,mtbr,bgong\}@google.com}
% For a paper whose authors are all at the same institution,
% omit the following lines up until the closing ``}''.
% Additional authors and addresses can be added with ``\and'',
% just like the second author.
% To save space, use either the email address or home page, not both
% \and
% Second Author\\
% Institution2\\
% First line of institution2 address\\
% {\tt\small secondauthor@i2.org}
}

\maketitle
%\thispagestyle{empty}

%%%%%%%%% ABSTRACT
\begin{abstract}
    Ensembling is a simple and popular technique for boosting evaluation performance by training multiple models (e.g., with different initializations) and aggregating their predictions. 
    This approach is commonly reserved for the largest models, as it is commonly held that increasing the model size provides a more substantial reduction in error than ensembling smaller models.
    However, we show results from experiments on CIFAR-10 and ImageNet that ensembles can outperform single models with both higher accuracy and requiring fewer total FLOPs to compute, even when those individual models' weights and hyperparameters are highly optimized.
    Furthermore, this gap in improvement widens as models become large.
    This presents an interesting observation that output diversity in ensembling can often be more efficient than training larger models, especially when the models approach the size of what their dataset can foster.
    Instead of using the common practice of tuning a single large model, one can use \textbf{ensembles as a more flexible trade-off between a model's inference speed and accuracy}.
    This also potentially eases hardware design, e.g., an easier way to parallelize the model across multiple workers for real-time or distributed inference.
\end{abstract}

\section{Introduction}
\label{sec:introduction}

Neural network ensembles are a popular technique to boost the performance of a model's metrics with minimal effort.
The most common approach in current literature involves training a neural architecture on the same dataset with different random initializations and averaging their output activations \cite{hinton2015distilling}.
This is known as {\it ensemble averaging}, or a simple type of {\it committee machine}.
For instance, on image classification on the ImageNet dataset, one can typically expect a 1-2\% top-1 accuracy improvement when ensembling two models this way, as demonstrated by AlexNet \cite{krizhevsky2012imagenet}.
Evidence suggests averaging ensembles works because each model will make some errors independent of one another due to the high variance inherent in neural networks with millions of parameters \cite{hansen1990neural, perrone1992networks, goodfellow2016deep}.

For ensembles with more than two models, accuracy can increase further, but with diminishing returns.
As such, this technique is typically used in the final stages of model tuning on the largest available model architectures to slightly increase the best evaluation metrics.
However, this method can be regarded as impractical for production use-cases that are under latency and size constraints, as it greatly increases computational cost for a modest reduction in error.

One may expect that increasing the number of parameters in a single network should result in higher evaluation performance than an ensemble of the same number of parameters or FLOPs, at least for models that do not overfit too heavily.
After all, the ensemble network will have less connectivity than the corresponding single network.
But we show cases where there is evidence to the contrary.
% \BG{Underscore this contrast to common believes in the abstract?}

In this paper, we show that we can consistently find averaged ensembles of networks with fewer FLOPs and yet higher accuracy than single models with the same underlying architecture.
This is true even for families of networks that are highly optimized in terms of its accuracy to FLOPs ratio.
We also show how this gap widens as the number of parameters and FLOPs increase.
We demonstrate this trend with a family of ResNets on CIFAR-10 \cite{zagoruyko2016wide} and EfficientNets on ImageNet \cite{tan2019efficientnet}.

The results of this finding imply that a large model, especially a model that is so large and begins to overfit to a dataset, can be replaced with an ensemble of a smaller version of the model for both higher accuracy and fewer FLOPs.
This can result in faster training and inference with minimal changes to an existing model architecture.
Moreover, as an added benefit, the individual models in the ensemble can be distributed to multiple workers which can speed up inference even more and potentially ease the design of specialized hardware.

Lastly, we experiment with this finding by varying the architectures of the models in ensemble averaging using neural architecture search to study if it can learn more diverse information associated with each model architecture.
Our experiments show that, surprisingly, we are unable to improve over the baseline approach of duplicating the same architecture in the ensemble in this manner.
Several factors could be attributed to this, including the choice of search space, architectural features, and reward function.
With this in mind, either more advanced methods are necessary to provide gains based on architecture, or it is the case that finding optimal single models would be more suitable for reducing errors and FLOPs than searching for different architectures in one ensemble.
% \BG{There could be multiple causes to this conclusion, such as search space, search methods, etc. Briefly explain them? Meanwhile, acknowledge and shed light on future work that could change what we have observed here.}

\section{Approaches and Experiments}
\label{sec:experiments}

For our experiments, we train and evaluate convolutional neural networks for image classification at various model sizes and ensemble them.
When ensembling, we train the same model architecture independently with random initializations, produce softmax predictions from each model, and calculate a geometric mean\footnote{Since the softmax applies a transformation in log-space, a geometric mean respects the relationship. We notice slightly improved ensemble accuracy when compared to an arithmetic mean.} $\mu$ across the model predictions. For $n$ models, we ensemble them by
\begin{equation}
    \mu = (y_1 y_2 \dots y_n)^{\frac{1}{n}}
\end{equation}
where the multiplication is element-wise for each prediction vector $y_i$.

We split our evaluation into two main experiments and a third follow-up experiment.

\subsection{Image Classification on CIFAR-10}

For the first experiment, we train wide residual networks on the CIFAR-10 dataset \cite{zagoruyko2016wide, krizhevsky2009learning}.
We train and evaluate the Wide ResNets at various width and depth scales to examine the relationship between classification accuracy and FLOPs and compare them with the ensembled versions of each of those models.
We train 8 models for each scale and ensemble them as described.
We select a depth parameter of $n = 16$, increase the model width scales $k \in \{1, 2, 4, 8\}$, and provide the corresponding FLOPs on images with a 32x32 resolution.
We use a standard training setup for each model as outlined in \cite{zagoruyko2016wide}.
% We also provide additional results in Appendix~\ref{sec:appendix} showing an increased depth parameter of $n = 20$ and varying the same width scales.

% \BG{Give more details about how you use the depth and scale parameters to slice the network? Add a citation for it if there is one?}

Note that we use smaller models than typically used (e.g., Wide ResNet 28-10) to show that our findings can work on smaller models that are less prone to overfitting.

\subsection{Image Classification on ImageNet}

To further show that the ensemble behavior as described can scale to larger datasets and more sophisticated models, we apply a similar experiment using EfficientNets on ImageNet \cite{tan2019efficientnet, russakovsky2015imagenet}.
EfficientNet provides a family of models using compound scaling on the network width, network depth, and image resolution, producing models from b0 to b7.
We adopt the first five of these for our experiments, training and ensembling up to three of the same model architecture on ImageNet and evaluating on the validation set.
We use the original training code and hyperparameters as provided by \cite{tan2019efficientnet} for each model size with no additional modifications.

\section{Results}
\label{sec:results}

In this section, we plot the relationship between accuracy and FLOPs for each ensembled model.
In cases of single models that are not ensembled, we plot the median accuracy.
We observe that the standard deviation of the evaluation accuracy of each model architecture size never exceeds 0.1\%, so we exclude it from the results for readability.
For models that are ensembled, we vary the number of $n$ trained models and choose the models randomly.
% Individual accuracies are tabulated in Appendix~\ref{sec:appendix}.

\begin{figure}[ht]
    \centering
    \includegraphics[width=\linewidth]{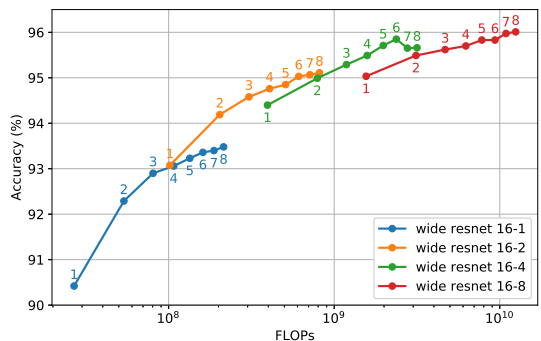}
    \caption{
        Test accuracy vs.\ model FLOPs (log-scale) when ensembling models trained on CIFAR-10.
        Each curve indicates the ensembles of increasing widths for a Wide ResNet $n_{d}$-$k$ with a depth of $n=16$.
        % The leftmost point plots the accuracy for a single model, while the rightmost point plots an ensemble of 8 of the same model.
        We show the number of models in each ensemble next to each point.
    }
    \label{fig:ensemble-comparison-cifar10}
\end{figure}

For the first experiment on CIFAR-10, Figure~\ref{fig:ensemble-comparison-cifar10} plots a comparison of Wide ResNets with a depth parameter of $n_{d}=16$ and width scales $k \in \{1, 2, 4, 8\}$.
For clarity in presentation, we show a smaller subset of all the networks we trained. For each network (e.g., ``wide restnet 16-8'', which stands for the depth parameter of $n_{d}=16$ and the width scale of $k=8$), we vary the number of models $n\in\{1,2,\cdots,8\}$ in an ensemble and label it alongside the curve.

% For a complete plot of all models (including varying depths) and additional training details, see Appendix~\ref{sec:appendix}.

\begin{figure}[ht]
    \centering
    \includegraphics[width=\linewidth]{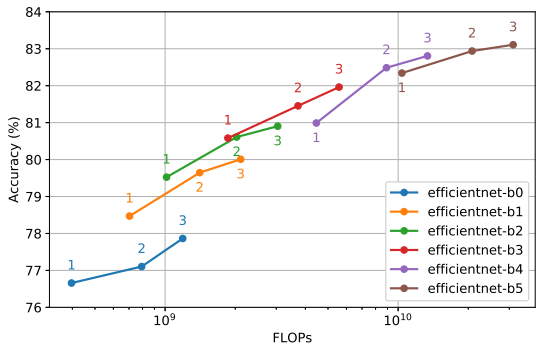}
    \caption{
        Validation accuracy vs.\ model FLOPs (log-scale) when ensembling models trained on ImageNet.
        Each curve indicates the ensembles of increasing sizes for a given EfficientNet.
        % The leftmost point plots the accuracy for a single model, while the rightmost point plots an ensemble of 3 of the same model.
        We show the number of models in each ensemble next to each point.
    }
    \label{fig:ensemble-comparison-imagenet}
    \vspace{-10pt}
\end{figure}

In the second experiment on ImageNet, Figure~\ref{fig:ensemble-comparison-imagenet} plots a comparison of EfficientNets b0 to b5.
%Note that the accuracy reported differ slightly from the original EfficientNet paper, as we re-train all models using the current EfficientNet code\footnote{The EfficientNet code can be found at \url{https://github.com/tensorflow/tpu/tree/master/models/official/efficientnet}} which may have changed since publication.We also do not use any image augmentation like AutoAugment or RandAugment to better observe the effects of overfitting.
Notably, we re-train all models using the current official EfficientNet code\footnote{The EfficientNet code can be found at \url{https://github.com/tensorflow/tpu/tree/master/models/official/efficientnet}}, but unlike the original paper that uses AutoAugment, here we do not use any specialized augmentation like AutoAugment or RandAugment to better observe the effects of overfitting.

\section{Discussion}

We draw the following observations from Figures~\ref{fig:ensemble-comparison-cifar10} and~\ref{fig:ensemble-comparison-imagenet} and particularly highlight the intriguing trade-off between accuracy and FLOPs thanks to the ensembling. 

First, we can see with no surprise that across the board, as the number of FLOPs increase for a single model, so too does the accuracy.
This is also true of the ensembles which essentially multiply the base FLOPs by $n$ for $n$ models.

What is more interesting is that the results show that there can be cases where \textbf{ensembles of models with fewer collective FLOPs can achieve higher accuracy than a single larger model}.
This is indicated by points that are above and to the left of other points.
For instance, an ensemble of eight Wide ResNet 16-2 models achieves the same accuracy of 95\% as a much wider Wide ResNet 16-8 at a fraction of the FLOPs (80M vs. 150M). An added benefit is that ensembles can easily be distributed to multiple workers to speed up computation even more.

Increasing the number of models in an ensemble will eventually be hit with diminishing returns, resulting a crossover point where an ensemble of the next largest model provides a better trade-off in terms of accuracy to FLOPs.
In CIFAR-10, we observe the optimal ensemble size would be 2-4 models before the accuracy improvement slows down.

Finally, an interesting trend is that for smaller models, we can see that ensembling them has a more difficult time improving over larger single models.
But as the models become larger, becoming increasingly likely to be over-parameterized and overfit to the dataset, we can see how ensembling provides a bigger accuracy boost over even larger models.
For instance, the ensembles of EfficientNet-b0 do not come close to reaching the same accuracy to FLOPs trade-off as EfficientNet-b1.
However, as the models become increasingly large, we see that the ensemble of two EfficientNet-b3 models achieves higher accuracy with fewer FLOPs over EfficientNet-b4, hence a better trade-off than EfficientNet-b4 provides.

Despite EfficientNet's scaling ability producing highly optimized models, we can still see gaps in performance where ensembles can perform better under the same number of total FLOPs, especially as the model size grows from b3 onwards. In other words, \textbf{ensembling offers an alternative and more effective scaling method than the compound scaling in EfficientNet} when some application scenarios permit the ensembling.

% For ImageNet, results are similar in nature but less pronounced. 
% Ensembles of smaller models do improve, but not at a higher accuracy to FLOPs ratio as seen in CIFAR-10.
% However, note that these models can be easily parallelized across multiple workers while achieving very similar accuracy.

% Despite EfficientNet being parameter and FLOP efficient, we begin to see that ensembling smaller models can beat out the next largest model, especially as the model size grows.

% Can Ensembling Ever Beat Similar Size Models in Classification Accuracy?
% Yes, but usually when the model is already over-parameterized.
% The plot shows crossover points where ensembles beat larger models with a similar number of parameters.
% Best results are achieved with 2-4 models before diminishing returns take over.

% For an ensemble that reaches accuracy/latency $a,l$, there likely exists a single model that can also reach accuracy no lower than $a$ and latency no higher than $l$.

\section{Neural Architecture Search (NAS) for Diverse Ensembles}

Having noted the observations above, we hypothesize that ensembles can be improved further by varying the architectures of each model in an ensemble rather than duplicating the same architecture.
The idea is that different architectures will naturally provide alternative features and therefore may enhance ensemble diversity.
This should, in turn, provide improved accuracy at no increase to the number of FLOPs.

\subsection{NAS Experiment Setup}
To test this hypothesis, we adopt the same NAS framework as MnasNet \cite{tan2019mnasnet}.
We use a search space predicting model depth, width, and convolution type.
We also augment the search space to include varying input resolution scales $r \in \{112, 168, 196, 224\}$.
As a result, each model provides $m = 50$ hyperparameters to search.
Additionally, we expand this to a joint search space to search for an ensemble of models by multiplying the search space $n$ times, one for each model, for a total of $nm$ hyperparameters.
Each model is trained individually and ensembled as described in earlier experiments.

We alter the reward function to be penalized by not the total latency of the ensemble, but the maximum latency of all of the models in an ensemble and simulate this latency on a Pixel Phone 1.
Assuming that each model can be run in parallel on separate workers, this would require the search to optimize the largest model in the ensemble at any given point to reduce the likelihood of producing ensembles where one model is large and the rest are anemic.
Lastly, we train each searched model for 10 epochs before evaluating the accuracy, which is part of the reward, on a held-out set.

% Reward is the product of the accuracy and the discounted latency of a searched model.
% Latency is simulated on a Pixel 1
% Hyperparameters are predicted using an actor-critic model with an RNN controller predicting a hyperparameter for each time step.
% During search, a model is trained for 10 epochs before evaluation.
% Use a joint search space over the ensemble of n models.
% Predict n*m hyperparameters.

\subsection{NAS Results}

\begin{figure}[ht]
    \centering
    \includegraphics[width=\linewidth]{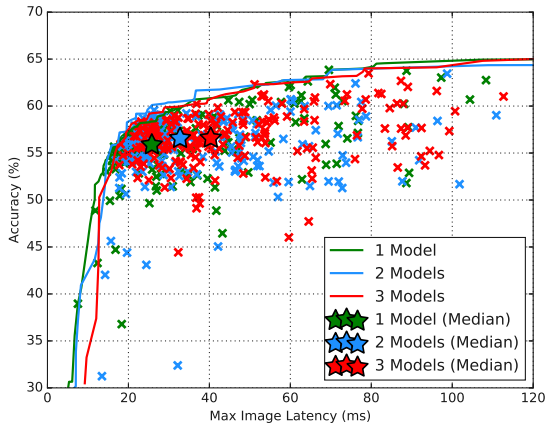}
    \caption{
        The resulting Pareto curves when searching for architectures across different ensemble sizes (trained for 10 epochs).
        We indicate the median ensemble accuracy and max latency of each ensemble size as stars in the plot.
    }
    \label{fig:multinas-pareto}
\end{figure}

We show the Pareto curves of the ensemble accuracy with respect to model maximum latency across ensembles of size one, two, and three in Figure~\ref{fig:multinas-pareto}.
This plot demonstrates the inherent trade-off between model accuracy and computation speed, with the best models being in the outer edge of the point cloud.

Results show that one-, two-, and three-model ensembles are surprisingly close to one another.
The skyline two-model ensembles tend to beat out single models, but only by 1\% at best.
Skyline three-model ensembles show nearly identical performance to single models.
We see that the median model accuracy does increase as the ensemble size grows, but at the cost of increased maximum latency.

Out of the  searched diverse models, we pick the most promising candidates for a target latency.
When trained to convergence, we find that two-model and three-model ensembles perform just as well as single models (assuming roughly equal max image latency). Somehow frustratingly,
we find that simply duplicating the best single model for a given latency target and ensembling them together provides the best improvement in accuracy.

This experiment presents evidence towards a conclusion that ensembles benefit the most from choosing the most accurate models and not models that are architecturally diverse, at least under our current NAS context.
For a fixed computational budget, this corresponds to using the best model architecture across the ensemble.
We of course caution that we only have tested this with a simple NAS setup on a single large image classification dataset.
This could change with a noisier and smaller dataset, or with more stringent constraints on model losses, regularization, or architectural mechanisms.

% Pareto Curves of rewards
% Curves are very similar to each other.
% 3 model ensembles show only 1\% improvement in accuracy over 1 model (in the first 10 epochs).

% Ensemble training can be very noisy.
% Prone to training divergence.
% Mixing pretrained weights with random weights leads to worse performance than with all random weights.
% There could be a disparity between the initial 10 epochs and final convergence that does not provide strong correlation for final accuracy.
% NAS can find very efficient, compact single models just as well as finding multiple ensembled models.
% In the case of image classification, it’s better to learn a single compact model rather than multiple models.
% It’s likely the task is not complex enough for over-parameterization to take effect for ensembles to provide a noticeable boost.

% Why the results for NAS are negative: models are too small, task is not noisy enough, models are too parameter/flop efficient.

% No significant difference between ensemble size and final accuracy

\section{Related Work}

Model ensembling has a long history with many different proposed techniques.
Most works in this area come before advancements in deep learning were popularized.
For instance, \cite{krogh1995neural} define different subsets of the training data and use cross-validation to divide data into different groups.
\cite{breiman1996bagging} developed bagging, where a different training set is given for different models to promote diversified feature learning.
And \cite{partridge1996engineering} is one of the earliest attempts at constructing ensembles with different models by changing the number of hidden nodes in each network.

% \cite{hinton2015distilling} (knowledge distillation). Using knowledge distillation for specialist models. ``learning specialist models that
% each focus on a different confusable subset of the classes can reduce the total amount of computation required to learn an ensemble.''

% \cite{opitz1996generating} (generating diverse ensembles. Provide a technique called Addemup that uses genetic algorithms to search for an accurate and diverse set of trained networks. Train on different partitions of the training set. Small changes in the dataset produce large changes in predictions.

% 

% 

% \cite{islam2003constructive} (constructive neural network ensembles). Adjust number of layers and number of networks simultaneously. Use negative correlation to train individual networks.

% 

% \cite{zhang2013exploiting} (Exploiting unlabeled data to enhance ensemble learning).

% \cite{siddique2019dynamic}. Add/delete hidden nodes of each network in an ensemble. Uses different training sets (bagging) and negative correlation loss.

% \cite{pang2019improving} (Improving Adversarial Robustness via Promoting Ensemble Diversity).

% \cite{sun2017ensemble} (ensembles for parallel training).

% \cite{asif2019ensemble} (Ensemble Knowledge Distillation for Learning Improved and Efficient Networks).

% Observation: most of these works use very noisy types of problems and datasets. Ensembles used here likely help smooth out the decision boundaries.

\section{Conclusion}

% \section{TODO Items}

% Training models individually vs. at the same time.
% Averaging before softmax vs. after.
% Arithmetic mean vs. geometric.
% Different activation functions.
% FLOPs vs. Parameters.
% Heterogeneous models, changing one architecture feature per model.

% Also begs the question: can ensembles beat models that are *maximally* parameter efficient?

% This also answers the question why the models in top leaderbords tend to be ensembles. At some point, a larger model will not offer anymore improvements, but it seems ensembling will nearly always improve no matter the type of task or the size of the model (as long as there are errors made independently).

We have demonstrated how averaging ensembles can result in higher accuracy with fewer FLOPs than popular single models on image classification.
This provides an interesting insight that smaller models can stand to provide great benefit without sacrificing on the accuracy to efficiency trade-offs of larger models. We advocate further inspections into the trade-off of ensembling especially for the applications where distributed inference is plausible.

{\small
\bibliographystyle{ieee_fullname}
\bibliography{egbib}
}

\end{document}